\documentclass[10pt,twocolumn,letterpaper]{article}

\usepackage[pagenumbers]{wacv} 

\usepackage{graphicx}
\usepackage{xcolor}
\usepackage{amsmath}
\usepackage{amssymb}
\usepackage{booktabs}
\usepackage{multirow}
\usepackage{amsmath}
\usepackage{enumitem}
\usepackage[T1]{fontenc}
\DeclareMathOperator*{\argmin}{argmin}
\usepackage[title]{appendix}

\usepackage[pagebackref,breaklinks,colorlinks]{hyperref}

\usepackage[capitalize]{cleveref}
\crefname{section}{Sec.}{Secs.}
\Crefname{section}{Section}{Sections}
\Crefname{table}{Table}{Tables}
\crefname{table}{Tab.}{Tabs.}
\Crefname{figure}{Figure}{Figures}
\crefname{figure}{Fig.}{Figs.}

\makeatletter
\robustify\@latex@warning@no@line
\makeatother
\usepackage[noblocks]{authblk}

\makeatletter
\renewcommand\AB@affilsepx{, \protect\Affilfont}
\makeatother
\newcommand{\appendixhead}%
{\centering\textbf{\huge Appendix}
\vspace{0.25in}}

\def\wacvPaperID{1672} 
\def\confName{WACV}
\def\confYear{2025}

\begin{document}

\title{
Exploring the Stability Gap in Continual Learning:\\The Role of the Classification Head
}

\author[1]{\vspace{-1cm}Wojciech~Łapacz\thanks{corresponding author, email: \href{mailto:wojciech.lapacz.stud@pw.edu.pl}{wojciech.lapacz.stud@pw.edu.pl}}}
\author[1,2]{Daniel~Marczak}
\author[1,2]{Filip~Szatkowski}
\author[1,2,3]{Tomasz~Trzciński}
\affil[1]{\normalsize Warsaw~University~of~Technology}
\affil[2]{IDEAS~NCBR}
\affil[3]{Tooploox}

\maketitle

\begin{abstract}

Continual learning (CL) has emerged as a critical area in machine learning, enabling neural networks to learn from evolving data distributions while mitigating catastrophic forgetting. However, recent research has identified the stability gap -- a phenomenon where models initially lose performance on previously learned tasks before partially recovering during training. Such learning dynamics are contradictory to the intuitive understanding of stability in continual learning where one would expect the performance to degrade gradually instead of rapidly decreasing and then partially recovering later. 
To better understand and alleviate the stability gap, we investigate it at different levels of the neural network architecture, particularly focusing on the role of the classification head. We introduce the nearest-mean classifier (NMC) as a tool to attribute the influence of the backbone and the classification head on the stability gap. 
Our experiments demonstrate that NMC not only improves final performance, but also significantly enhances training stability across various continual learning benchmarks, including CIFAR100, ImageNet100, CUB-200, and FGVC Aircrafts. Moreover, we find that NMC also reduces task-recency bias. Our analysis provides new insights into the stability gap and suggests that the primary contributor to this phenomenon is the linear head, rather than the insufficient representation learning.
\end{abstract}

\section{Introduction}
\label{sec:introduction}

Neural networks have been widely adopted across various domains, including computer vision, natural language processing, speech and audio processing, and control tasks~\cite{redmon2016lookonceunifiedrealtime, hannun2014deepspeechscalingendtoend, caruana2015intelligible}. However, most neural network applications are limited to offline settings with static data distributions, primarily due to the challenge of catastrophic forgetting~\cite{mccloskey1989catastrophic}—a phenomenon where a model loses previously acquired knowledge when exposed to new data or distribution shifts. Adapting to and learning from evolving data distributions is crucial for many tasks, such as autonomous driving~\cite{shaheen2022continual}, robotics~\cite{lesort2020continual}, edge devices~\cite{pellegrini2021continual}, and reinforcement learning~\cite{wolczyk2021continual}. Consequently, the field of continual learning (CL)~\cite{hadsell2020embracing,van2019three,masana2022class} has emerged to enable neural networks to learn from dynamic data streams while mitigating catastrophic forgetting.

\begin{figure}[!t]
  \centering
  \includegraphics[width=1.0\linewidth]{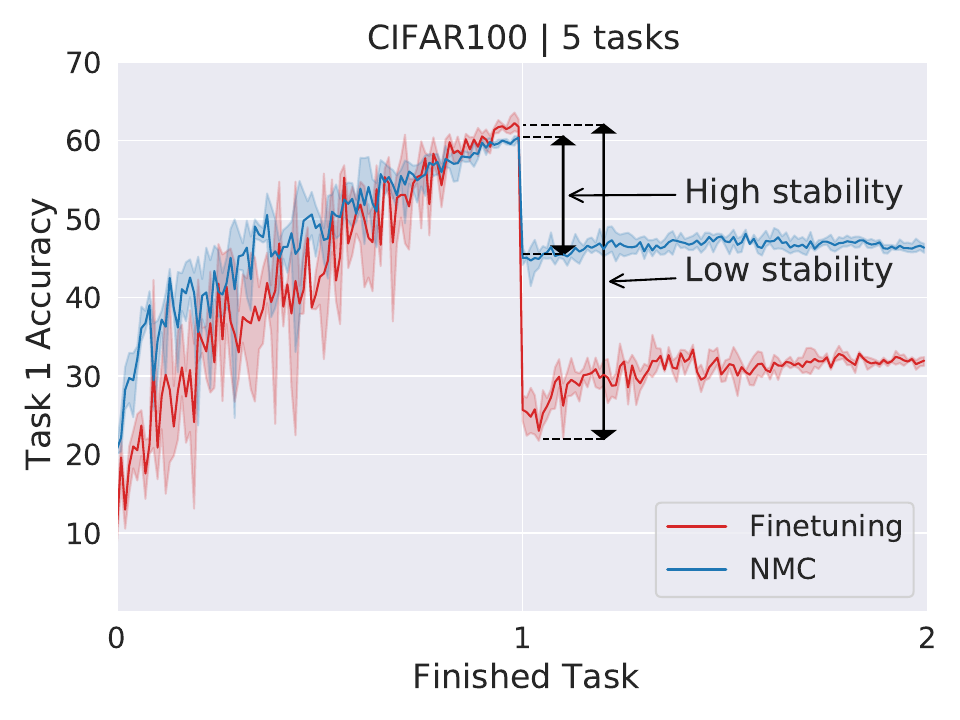}
   \caption{
   Stability gap phenomenon throughout learning the first two tasks from the CIFAR100 5-task split. When evaluated with a standard linear head and Nearest-Mean Classifier (NMC), the NMC performance on the first task with more stable through the learning phase and achieves a better final accuracy, even though both networks use the same representations.
   }
   \label{fig:teaser}
\end{figure}


Continual learning methods developed over the years mitigate forgetting and facilitate good performance on continual learning benchmarks through several techniques such as regularization~\cite{kirkpatrick2017overcoming,li2017learning}, replay~\cite{riemer2018learning,chaudhry2019tiny,shin2017continual,buzzega2020dark}, parameter isolation~\cite{rusu2016progressive,li2019learn,von2019continual,wang2022dualprompt} and their hybrid combinations~\cite{rebuffi2017icarl,hou2019learning,wu2019large,ahn2021ss}.
Despite recent progress in the field~\cite{masana2022class}, usually measured by the final performance of the continual learner on all tasks, training dynamics remain relatively underexplored.
Recent work of Lange et al.~\cite{lange2022continual} sheds more light on how the network behaves during continual training and highlights a curious phenomenon of the stability gap. The authors show that CL methods evaluated throughout the learning on the new task initially forget the previously learned knowledge and only recover a part of the performance on the previously learned tasks in the later stages of training. The stability gap brings into question a traditional understanding of stability in continual learning. In light of the stability gap, it appears that instead of mitigating actual forgetting, continual learning methods allow for initial forgetting to facilitate knowledge recovery later. Further investigation of this phenomenon is required to understand continual learning dynamics better.

Initial work on the stability gap focused on image classification and class-, task- and domain-incremental continual learning settings~\cite{lange2022continual}. Later works demonstrated that the stability gap also occurs during continual pre-training of Large Language Models~\cite{guo2024efficient}, and even incremental training with full replay~\cite{hess2023two,kamath2024expanding}. The fact that the stability gap occurs even without a task or domain shift suggests that it might be an inherent property of the optimization algorithms used to train neural networks. Nevertheless, continual learning is the domain most affected by its presence, as its algorithms are particularly exposed to significant data shifts and are often required to function in an online manner. Therefore, recently a few works started to explore mitigating the stability gap in continual learning~\cite{carta2023improving,harun2023overcoming}, so far within a limited scope. To mitigate the impact of the stability gap across various applications of continual learning, we still need a better understanding of this phenomenon.

Previous works~\cite{ramasesh2020anatomy} discovered that higher layers of neural networks are responsible for most of the forgetting that occurs in the continually trained neural network. That raises the question whether we can locate the source of the stability gap. Therefore, we pose the following research question:


\begin{center}
\textit{\textbf{Which part of the neural network\\is the primary source of the stability gap?}}
\end{center}

To this end, we disentangle the problem into backbone- and classifier-level stability gap by employing nearest-mean classifier (NMC) - a powerful continual learning tool~\cite{rebuffi2017icarl,marczak2023generalized,goswami2024fecam} that, to the best of our knowledge, has been overlooked in previous works addressing the stability gap. NMC classifies the samples based on their proximity to the class prototypes in the representation space and therefore offers a disentangled classification framework that we can compare with a classic linear head learned with cross-entropy. We show that an oracle NMC (which uses prototypes constructed from all training samples seen so far) can mitigate most of the instability compared to using a linear head.

We show that even in a more realistic scenario with a limited number of exemplars, an NMC classifier on top of representations trained with cross-entropy not only outperforms a linear head, but also exhibits way higher stability through the training, mitigating a large part of the stability gap when compared to the standard approach (see \cref{fig:teaser}). We provide a thorough evaluation of NMC in comparison with the standard replay approach on well-established continual learning benchmarks split CIFAR100, ImageNet100 and fine-grained classification datasets CUB-200 and FGVC Aircrafts. We show that in all cases NMC facilitates more stable training and improves the final performance, regardless of the size of the memory. In addition, we demonstrate that the NMC classifier also alleviates a large part of task-recency bias\cite{wu2019large,ahn2021ss} - networks' preference towards predicting the classes from the latest task.

\noindent The main contributions of our paper can be summarised as: 
\begin{itemize} 
\item{We are the first to investigate the stability gap in the context of different parts of the network.}
\item{We compare the performance and stability of linear head and NMC across a wide range of well-established continual learning baselines.} 
\item{To strengthen our findings, we provide a thorough analysis of the phenomena described above examining the impact of memory buffer size, initialization with the pre-trained network, and task-recency bias exhibited by the final model.}
\end{itemize}
Our findings indicate that \textbf{the linear head is responsible for the main portion of the stability gap phenomenon, and that the network representations are sufficient for way more stability when used with non-parametric NMC classifier}. We hope that our work will shed new light on the stability gap in continual learning and prompt further investigation of this phenomenon.

\section{Related Works}
\label{sec:relate}

\subsection{Continual Learning}
Continual learning aims to enable learning from changing data streams and mitigate catastrophic forgetting~\cite{mccloskey1989catastrophic} - destructive overwriting of previously acquired knowledge that appears when learning from new data. Continual learning usually assumes that the data come in the form of sequential tasks, and once the learner finishes processing the task it can no longer access its data for training; however, it is still evaluated on the data from all the tasks seen so far. The main scenarios analyzed in the continual learning community are task-incremental learning, class-incremental learning, and domain-incremental learning~\cite{van2019three,masana2022class}. Task-incremental learning and class-incremental learning usually analyze a scenario where new tasks contain new classes, with task-incremental learning assuming access to task identity information during prediction and class-incremental learning assuming no access to such information, which makes it a more challenging scenario. Domain-incremental learning instead operates on the set of the same classes but with the data distribution changing between tasks (e.g. changing camera, lighting, or image style). Continual learning approaches are usually categorized as regularization-based, replay-based or architecture-based, with many works also proposing hybrid approaches~\cite{masana2022class}. Regularization-based methods try to prevent forgetting by penalizing changes to model weights~\cite{kirkpatrick2017overcoming} or activations~\cite{li2017learning}. Replay- or exemplar-based methods enforce the stability of the network by including a small subset of old data during training on a new task to rehearse previously learned knowledge; the exemplars can be either stored samples from old tasks~\cite{riemer2018learning,chaudhry2019tiny}, samples generated by a generative model~\cite{shin2017continual} or stored features from the old data~\cite{buzzega2020dark}. Architectural methods, or parameter isolation methods, try to alleviate forgetting by allocating specific network parameters to specific tasks or growing the network while training~\cite{rusu2016progressive,li2019learn,von2019continual,wang2022dualprompt}. Most of the more modern methods try to combine the strengths of multiple approaches and propose hybrid solutions~\cite{rebuffi2017icarl,hou2019learning,wu2019large,ahn2021ss}

\subsection{Stability Gap}
Stability gap~\cite{lange2022continual} is a phenomenon where a continually trained neural network experiences a significant forgetting on task transitions followed by a recovery phase in which the network recovers a significant fraction of its original performance.
The pioneering work~\cite{lange2022continual} identifies the stability gap in class-, task- and domain-incremental settings for image classification, while~\cite{guo2024efficient} identifies similar phenomenon in continual pre-training of Large Language Models (LLMs).
\cite{hess2023two} shows that the stability gap occurs even with incremental joint training (‘full replay’). \cite{kamath2024expanding} goes a step further, demonstrating that it is also present in joint incremental learning of homogeneous tasks (all tasks are drawn from the same distribution). 
Carta et al.~\cite{carta2023improving} try to overcome the stability gap and improve the network stability with temporal ensembling, but the scope of their work is limited to online continual learning. Harun and Kanan~\cite{harun2023overcoming} also investigate the stability gap in a setting resembling out-of-distribution (OOD) detection and propose a method that combines multiple heuristics to address it. However, to the best of our knowledge, no prior work has studied the disentangled influence of the backbone and classification heads on the stability gap.

\subsection{Nearest Mean Classifier (NMC) in CL}
NMC is an alternative to a linear classifier that utilizes the prototypes (average representations of each class based on a set of samples) for classifying new samples based on the distance between their representations and the prototypes. NMC usually assigns the class of the closest prototype and as such is a non-parametric classifier that instead of weights requires storing class prototypes. NMC and its variants have been widely used in few-shot or zero-shot learning \cite{wang2020fewshot, xian2020zeroshot}. The first major application of NMC in continual learning was iCaRL~\cite{rebuffi2017icarl}, which classifies old classes by calculating class prototypes from the exemplars stored in the memory buffer. Since then, many methods~\cite{mai2021supervised,goswami2024fecam} employed NMC to improve continual learning performance.
Several works also explored exemplar selection strategies for NMC~\cite{rebuffi2017icarl,broderic2013streamvarbay}, but without significant improvements over simple random selection. So far, no work has explored the phenomenon of the stability gap in the context of the nearest-mean classifier (NMC). 

\section{Problem Setup}
\label{setup}

\subsection{Incremental learning}
\label{sec:incr-learn}
We consider a continual learning scenario with a learner $f$, a neural network composed of the backbone $\Theta$, and a linear classification head $g$. Given input $x$, the model produces the prediction by applying the classifier on top of the features extracted from the backbone: $f(x) = g(\Theta(x))$.

The learner is trained on a sequence of $T$ tasks, which correspond to separate, disjoint datasets $D_1, ..., D_{T}$. We refer to the model obtained after $t$ tasks as $f_t$. For task $t$, $D_t$ contains inputs $x_t$ and corresponding ground truth labels $y_t$. The datasets are presented sequentially in an offline manner, meaning that the model can pass through the training data multiple times. During training on task $t$, the model can no longer access the full data corresponding to the previous tasks $t-1, t-2, ..., 1$. This differs from a scenario called joint incremental learning, where the model can continuously train on the union of all the datasets seen so far. Joint incremental learning is considered the upper bound on continual learning performance~\cite{masana2022class}. In practice, many continual learning methods store a small number of exemplars to rehearse previously learned knowledge, as this guarantees some degree of continual performance.

We consider a standard class-incremental scenario~\cite{masana2022class,van2019three}, where each task contains new classes previously unseen by the model. Upon encountering a new task $t$, we create a dedicated task head $g_t$ that will predict the set of classes in this task. However, as the model cannot access the task identity in the class-incremental scenario, we use the concatenated output of heads for all the tasks seen so far to obtain the prediction.

We employ the Nearest-Mean Classifier (NMC) as introduced in iCaRL~\cite{rebuffi2017icarl}. To predict label $y^*$, we compute prototype vectors $\mu_1, ..., \mu_C$ (mean features) of all $C$ classes seen so far using exemplars for previous tasks classes, and training set for current task classes. Then, we compute distances between the new feature vector and all prototypes, assigning the class label of the most similar prototype:

\begin{equation}
    y^* = \argmin_{c=1,...,C} ||f(x) - \mu_c||
    \label{eq:nmc}
\end{equation}

To ensure a fair comparison between finetuning and NMC, \textbf{in all experiments we train both models in the same manner and only change the classification algorithm during the prediction.}

\subsection{Metrics}
\label{sec:metrics}
In this section, we describe the metrics we use to evaluate the performance of continual learners. We employ the standard average accuracy metric defined by Masana et al.~\cite{masana2022class} and average minimum accuracy and worst-case accuracy from the initial stability gap paper~\cite{lange2022continual}. Finally, we define the stability gap as the normalized difference between the model performance at the end of the previous task and the lowest accuracy obtained through model learning on the current task. All metrics here are defined based on Task-Agnostic Accuracy, where the model does not have information about task ID. Additional study using Task-Aware Accuracy is presented in the Appendix. For the reader's convenience, we describe those metrics shortly below.

\noindent\textbf{Average Accuracy.} We define average incremental accuracy $\textbf{ACC}_{t}$ obtained after task $t$ as:
\begin{equation}
    \textbf{ACC}_{t} = \frac{1}{t}\sum_{i=1}^{t} \textbf{A}(f_{t}, y_i)
    \label{eq:avg-acc}
\end{equation}
where ${\textbf{A}(f_t, y_i)}$ is the accuracy of the model $f_t$ on the $i$-th task data, with $i \leq t$.

\noindent\textbf{Average minimum accuracy.} We define average absolute minimum accuracy $\textbf{min-ACC}_{t}$ measured over previous evaluation tasks after they have been learned as:
\begin{equation}
    \label{eq:min-acc}
    \textbf{min-ACC}_{t} =  \frac{1}{t-1}\sum_{i=1}^{t-1} \min_n \textbf{A}(f_{n}, y_{i}), 
\end{equation}
where $i < n \leq t$. Intuitively, this metric gives
a worst-case estimation of knowledge preservation on the previously observed tasks.

\noindent\textbf{Worst-case accuracy.} We define $\textbf{WC-ACC}_{t,k}$ as a trade-off between the accuracy on $k$-th iteration of task $t$ and $\textbf{min-ACC}_{t}$ for previous tasks:
\begin{equation}
    \label{eq:wc-acc}
    \textbf{WC-ACC}_{t,k} = \frac{1}{t} \ \textbf{A}(f_k, y_t) + (1-\frac{1}{t}) \ \textbf{min-ACC}_{t}
\end{equation}
 
\noindent\textbf{Stability gap.} We introduce stability gap $\textbf{SG}_{t, i}$, a metric to measure that we define as a maximum drop in accuracy on the task $i < t$ throughout learning a new task t: 
\begin{equation}
    \textbf{SG}_{t, i} = \frac{\textbf{A}(f_{t-1}, y_i) - \min_n \textbf{A}(f_{t}^n, y_i)}{\textbf{A}(f_{t-1}, y_i)}.
    \label{eq:sg}
\end{equation}
where $f_t^n$ refers to a model obtained after $n$ epochs on a task $t$. This metric gives us an insight into how much the model degrades during training. However, note that it does not take into account final accuracy, and a learner whose performance initially drops to zero and then recovers all its accuracy on an old task throughout the training on a new task will still exhibit a high stability gap. To properly measure and compare the stability gap in cases where the initial model already exhibits very low accuracy, we normalize the metric with the starting accuracy.

\section{Experiments}
\label{sec:experiments}

\begin{figure*}[!t]
  \centering
    \includegraphics[width=1.0\linewidth]{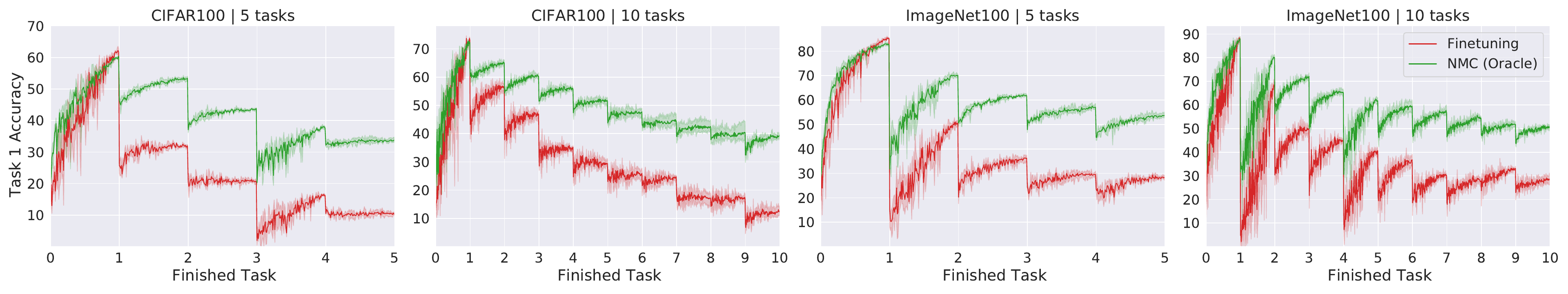}
  \caption{Oracle NMC leads to significantly better CL performance than using classification head, exhibiting lower performance drops on task boundaries and higher final accuracy. We report first task accuracy in \%.}
  \label{fig:disentangling}
\end{figure*}


\subsection{Setup}
We conduct experiments on common continual learning benchmarks CIFAR100~\cite{Krizhevsky2009LearningML},
ImageNet100~\cite{tian2019contrastive} and fine-grained classification datasets CUB-200-2011 Birds (Birds)~\cite{wah2011caltech} and FGVC Aircraft (Aircrafts)~\cite{maji2013fine}. We create CIL scenarios by splitting the classes in each dataset into equally sized disjoint tasks.

We use the code built on top of~FACIL~\cite{masana2022class} framework. We use ResNet18~\cite{he2016resnet} as a backbone and train from random initialization (except \cref{sec:finegrained-from-pretrained} where we start from the weights pre-trained on ImageNet). In the Appendix, we provide the ablation study of different CNN architectures used as backbones. We use the same hyperparameters for all variants within the single method unless stated otherwise and report our results averaged over three runs with different random seeds.

In every setup, we train the network on each new task for 100 epochs with a batch size of 128. We use an SGD optimizer with a linearly decaying learning rate. For the CIFAR100 and ImageNet100 datasets, we utilize a constant memory budget of 2000 exemplars. This approach ensures a balanced representation of past tasks while limiting the total memory usage. For experiments on fine-grained datasets, we use 10 exemplars per class. Exemplars are selected randomly at the end of each task.
We compute all the metrics from \cref{sec:metrics} for the full test set at the end of each epoch.

\subsection{Disentangling stability gap}
\label{sec:disentangle}

To understand which part of the network contributes the most to the stability gap and further disentangle the performance of the classification head from the backbone, we consider an \textit{oracle NMC}. It is trained in the same manner as fine-tuning (with exemplars) and NMC (on data from current task and from the memory buffer) but additionally it has access to all the training data seen so far when computing the prototypes. This setup gives us the upper bound on the non-parametric classification based on the network representations. By comparing the difference between the performance of the oracle NMC and the performance of the network using a linear head, we can isolate the portion of the stability gap that can be attributed to the classifier.

We present the results in \cref{fig:disentangling} and observe that the oracle NMC greatly surpasses standard finetuning. Specifically, the numerical results in \cref{tab:disentangling} reveal that we can attribute most of the stability gap to the linear head, with the gap being 2 to 4 times greater for the linear head network. The remaining portion of the stability gap is an upper bound on the gap caused by the backbone. Our findings confirm our hypothesis that the primary cause of the stability gap lies in the classification head, rather than in the quality of the network's representations.

Results presented in \cref{fig:disentangling,tab:disentangling}, however, are based on the oracle NMC which relies on access to all the data seen so far while building a non-parametric classifier. They provide valuable insight into the crucial (negative) impact of classification head on the stability in CL. However, they cannot be used directly in practical scenarios with limited memory buffers.

\begin{table}[t]
    \centering
    \caption{Stability metrics for standard finetuning compared with oracle NMC that computes prototypes based on the whole training set. \textbf{The gap between oracle NMC and finetuning can be attributed to the linear head, as both cases use the same backbone representations.} We report all the metrics at the end of the sequential training in \%.}
    \scalebox{0.75}{
    \begin{tabular}{lcccc}
    \toprule
    & WC-ACC ($\uparrow$) & min-ACC ($\uparrow$) & SG($\downarrow$) & ACC ($\uparrow$) \\ \midrule
    & \multicolumn{4}{c}{CIFAR100/5} \\
    \cmidrule(lr){2-5} 
    FT                                          & 23.12\scriptsize{$\pm$0.11} & 11.9\scriptsize{$\pm$0.20} &  54.02\scriptsize{$\pm$0.99} & 25.52\scriptsize{$\pm$0.42}    \\
    +NMC Oracle                                   & \textbf{36.35}\scriptsize{$\pm$\textbf{0.49}} & \textbf{35.48}\scriptsize{$\pm$\textbf{0.46}} & \textbf{13.88}\scriptsize{$\pm$\textbf{1.22}} & \textbf{38.76}\scriptsize{$\pm$\textbf{0.65}}   \\
    \midrule
     & \multicolumn{4}{c}{CIFAR100/10} \\
    \cmidrule(lr){2-5} 
    FT                                          & 13.94\scriptsize{$\pm$0.39} & 7.58\scriptsize{$\pm$0.32} & 63.04\scriptsize{$\pm$2.57} & 21.04\scriptsize{$\pm$0.31}   \\
    +NMC Oracle                                    & \textbf{29.87}\scriptsize{$\pm$\textbf{0.17}} & \textbf{28.82}\scriptsize{$\pm$\textbf{0.22}}  & \textbf{22.76}\scriptsize{$\pm$\textbf{1.40}} & \textbf{36.12}\scriptsize{$\pm$\textbf{0.40}}   \\ 
        \midrule
    & \multicolumn{4}{c}{ImageNet100/5} \\
    \cmidrule(lr){2-5} 
    FT                                          &  29.57\scriptsize{$\pm$0.62} & 15.9\scriptsize{$\pm$0.72} & 61.62\scriptsize{$\pm$2.00} & 40.50\scriptsize{$\pm$0.48}  \\
    +NMC Oracle                                    &  \textbf{48.43}\scriptsize{$\pm$\textbf{1.94}} & \textbf{44.1}\scriptsize{$\pm$\textbf{2.12}} & \textbf{28.77}\scriptsize{$\pm$\textbf{2.95}} & \textbf{57.36}\scriptsize{$\pm$\textbf{0.61}}  \\
    \midrule
        & \multicolumn{4}{c}{ImageNet100/10} \\
    \cmidrule(lr){2-5} 
    FT                                          & 24.8\scriptsize{$\pm$1.13}   &   17.45\scriptsize{$\pm$1.21}    & 49.20\scriptsize{$\pm$2.70}  &   32.38\scriptsize{$\pm$0.37}  \\
    +NMC Oracle                                    &  \textbf{45.18}\scriptsize{$\pm$\textbf{0.49}}   &   \textbf{44.71}\scriptsize{$\pm$\textbf{0.56}}   & \textbf{12.07}\scriptsize{$\pm$\textbf{1.64}} &   \textbf{51.17}\scriptsize{$\pm$\textbf{0.59}}  \\ 
    \bottomrule
    \end{tabular}
    }
    \label{tab:disentangling}
    \vspace{-0.3cm}
\end{table}

\subsection{Realistic NMC scenarios}
\label{sec:realistic-scenarios}

\begin{figure*}[!t]
  \centering
    \includegraphics[width=1.0\linewidth]{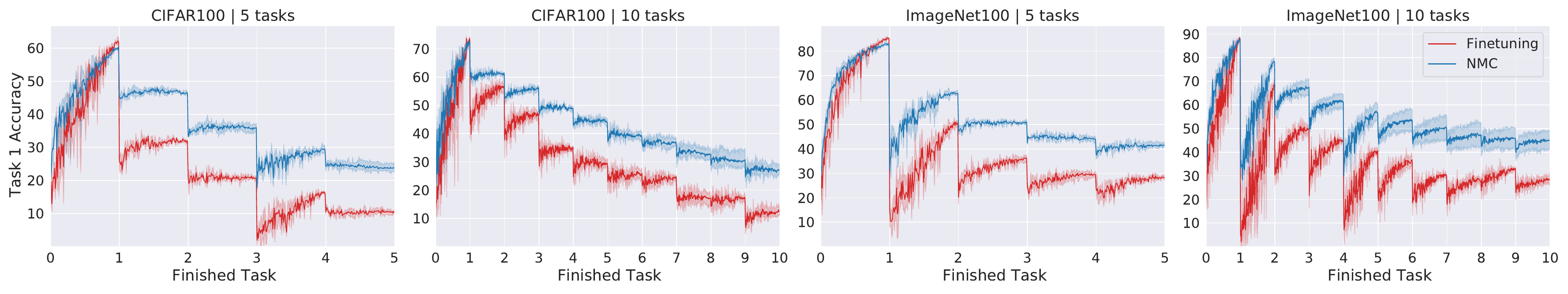}
  \caption{First task accuracy (\%) for NMC and linear head on standard continual learning benchmarks. NMC shows higher performance and stability through the training across all the evaluated benchmarks.}
  \label{fig:main}
\end{figure*}


\begin{figure*}[t]
  \centering
    \includegraphics[width=1.0\linewidth]{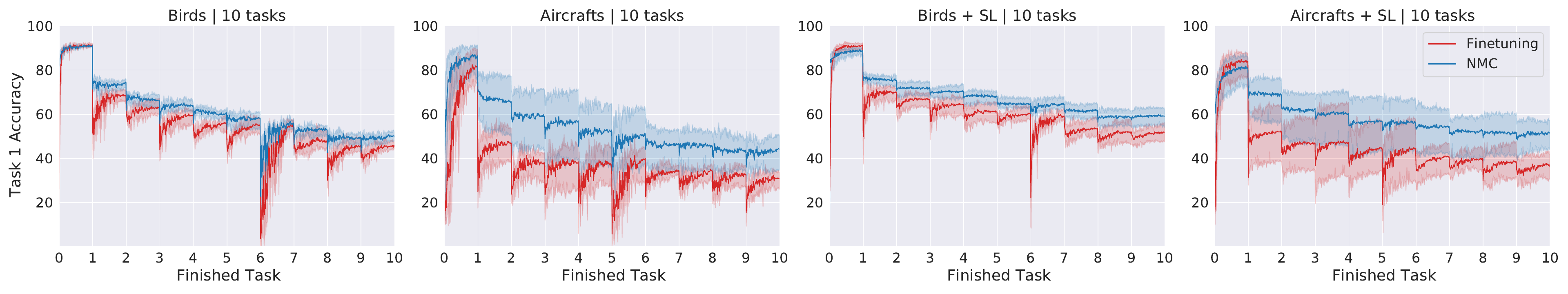}
    \caption{First task accuracy (\%) for regular training (left) and Slow Learner (right) on fine-grained classification benchmarks when starting from pre-trained model. NMC enables better stability in such a setting as well.}
\label{fig:finegrained}
\end{figure*}



\begin{table}[t]
    \centering
    \caption{
    NMC improves the stability and final accuracy across general image classification benchmarks. We report all the metrics at the end of the sequential training in \%.
    }
    \scalebox{0.8}{
    \begin{tabular}{lcccc}
    \toprule
    & WC-ACC ($\uparrow$) & min-ACC ($\uparrow$) & SG($\downarrow$) & ACC ($\uparrow$) \\ \midrule
    & \multicolumn{4}{c}{CIFAR100/5} \\
    \cmidrule(lr){2-5}
    FT                                          & 23.12\scriptsize{$\pm$0.11} & 11.9\scriptsize{$\pm$0.20} &  54.02\scriptsize{$\pm$0.99} & 25.52\scriptsize{$\pm$0.42}    \\
    +NMC                                         & \textbf{34.29}\scriptsize{$\pm$\textbf{0.60}} & \textbf{29.38}\scriptsize{$\pm$\textbf{0.68}} & \textbf{23.85}\scriptsize{$\pm$\textbf{0.88}} & \textbf{35.24}\scriptsize{$\pm$\textbf{0.78}}   \\
    \midrule
     & \multicolumn{4}{c}{CIFAR100/10} \\
    \cmidrule(lr){2-5}
    FT                                          & 13.94\scriptsize{$\pm$0.39} & 7.58\scriptsize{$\pm$0.32} & 63.04\scriptsize{$\pm$2.57} & 21.04\scriptsize{$\pm$0.31}   \\
    +NMC                                         & \textbf{27.27}\scriptsize{$\pm$\textbf{0.50}} & \textbf{24.01}\scriptsize{$\pm$\textbf{0.65}} & \textbf{25.35}\scriptsize{$\pm$\textbf{2.66}} & \textbf{31.14}\scriptsize{$\pm$\textbf{0.38}}  \\ \midrule
     & \multicolumn{4}{c}{ImageNet100/5} \\
    \cmidrule(lr){2-5}
    FT                                          &  29.57\scriptsize{$\pm$0.62} & 15.9\scriptsize{$\pm$0.72} & 61.62\scriptsize{$\pm$2.00} & 40.50\scriptsize{$\pm$0.48}  \\
    +NMC                                         &  \textbf{47.27}\scriptsize{$\pm$\textbf{1.43}} & \textbf{39.24}\scriptsize{$\pm$\textbf{1.60}} & \textbf{29.85}\scriptsize{$\pm$\textbf{1.84}} & \textbf{52.15}\scriptsize{$\pm$\textbf{0.75}}  \\
    \midrule
     & \multicolumn{4}{c}{ImageNet100/10} \\
    \cmidrule(lr){2-5}
    FT                                          & 24.8\scriptsize{$\pm$1.13}   &   17.45\scriptsize{$\pm$1.21}    & 49.20\scriptsize{$\pm$2.70}  &   32.38\scriptsize{$\pm$0.37}  \\
    +NMC                                         &  \textbf{41.98}\scriptsize{$\pm$\textbf{0.88}}   &   \textbf{37.26}\scriptsize{$\pm$\textbf{0.83}}  & \textbf{20.03}\scriptsize{$\pm$\textbf{1.84}}  &   \textbf{45.02}\scriptsize{$\pm$\textbf{0.77}}  \\
    \bottomrule
    \end{tabular}}
    \label{tab:main-results}
    \vspace{-0.3cm}
\end{table}

We build on the insights from the previous Section and utilize NMC's positive impact on stability in more realistic scenarios of limited memory buffer. We provide additional analysis on other CIL approaches in the Appendix.

\noindent\textbf{Classic CL benchmarks}
We present our main results in~\cref{tab:main-results} and~\cref{fig:main}. We observe that NMC significantly improves the stability of a continually trained network even when the prototypes are calculated only on the samples from the memory buffer. The difference in stability and final performance of these two approaches shows that the classification head is responsible for a large fraction of the instabilities and performance loss during the continual training.

\begin{table}[t]
    \centering
    \caption{
    NMC improves the stability when of finetuning pretrained models on the sequences of fine-grained datasets. Moreover, Slow Learner (SL) further improves the results of both FT and NMC and the advantage of NMC over FT remains. We report all the metrics at the end of the sequential training in \%.
    }
    \label{tab:finegrained}
    \scalebox{0.8}{
    \begin{tabular}{lcccc}
    \toprule
    & WC-ACC ($\uparrow$) & min-ACC ($\uparrow$) & SG($\downarrow$) & ACC ($\uparrow$) \\ \midrule
     & \multicolumn{4}{c}{Birds/10} \\ \cmidrule{2-5} 
    FT & 49.24\scriptsize{$\pm$0.27} & 45.45\scriptsize{$\pm$0.65} & 20.94\scriptsize{$\pm$1.56} & 56.02\scriptsize{$\pm$0.55}   \\
    +NMC & \textbf{53.89}\scriptsize{$\pm$\textbf{0.66}} & \textbf{51.06}\scriptsize{$\pm$\textbf{0.80}} & \textbf{14.74}\scriptsize{$\pm$\textbf{0.50}} & \textbf{58.39}\scriptsize{$\pm$\textbf{0.52}}    \\ 
    \midrule
    FT + SL                                         & 57.36\scriptsize{$\pm$0.83} & 54.38\scriptsize{$\pm$1.58} & 12.95\scriptsize{$\pm$3.57} & 60.91\scriptsize{$\pm$0.50}     \\
    +NMC                                        & \textbf{61.28}\scriptsize{$\pm$\textbf{0.33}} & \textbf{60.08}\scriptsize{$\pm$\textbf{0.89}} & \textbf{7.00}\scriptsize{$\pm$\textbf{0.86}} & \textbf{62.86}\scriptsize{$\pm$\textbf{0.43}}    \\ \midrule
            & \multicolumn{4}{c}{Aircrafts/10} \\ \cmidrule{2-5}
    FT &   24.83\scriptsize{$\pm$2.89}   &   17.32\scriptsize{$\pm$2.91}   & 62.22\scriptsize{$\pm$3.45}  &   44.60\scriptsize{$\pm$1.74}  \\
    +NMC &   \textbf{46.72}\scriptsize{$\pm$\textbf{2.97}}   &   \textbf{42.12}\scriptsize{$\pm$\textbf{2.99}}   & \textbf{22.98}\scriptsize{$\pm$\textbf{4.44}}  &   \textbf{54.25}\scriptsize{$\pm$\textbf{0.05}}  \\  \midrule
    FT + SL                                         &  41.91\scriptsize{$\pm$1.82}   &   36.26\scriptsize{$\pm$2.11} & 26.96\scriptsize{$\pm$3.25}  &   49.20\scriptsize{$\pm$0.45}  \\
    +NMC                                        &  \textbf{53.80}\scriptsize{$\pm$\textbf{0.87}}   &   \textbf{51.58}\scriptsize{$\pm$\textbf{1.59}}   & \textbf{10.38}\scriptsize{$\pm$\textbf{1.19}}  &   \textbf{56.52}\scriptsize{$\pm$\textbf{0.85}}  \\
    \bottomrule
    \end{tabular}
    }
    \vspace{-0.3cm}
\end{table}

\noindent\textbf{Fine-grained tasks with pre-trained model}
\label{sec:finegrained-from-pretrained}
In this Section, we evaluate if our observations hold in settings where we start from a pre-trained model (ResNet-18 backbone pre-trained on ImageNet). We selected fine-grained datasets to create task sequences that remain challenging for the backbone pre-trained on ImageNet. We present the results in ~\cref{tab:finegrained} and~\cref{fig:finegrained}. We observe that NMC significantly improves the stability metrics and the final performance of the model, \textit{e.g.} on Aircrafts it improves stability almost threefold (looking at SG metric) and improves the final accuracy by almost 10 p.p. Moreover, we experiment with using Slow Learner~\cite{zhang2023slca} -- a technique of continually training pre-trained models that employs a lower learning rate for the backbone (in our case 0.01) and higher learning rate for classification rate (in our case 0.1). We observe that SL improves the stability and the final accuracy. However, even when combined with SL, NMC retains its edge over FT.  

\begin{figure*}[!ht]
    \centering
    \includegraphics[width=1.0\linewidth]{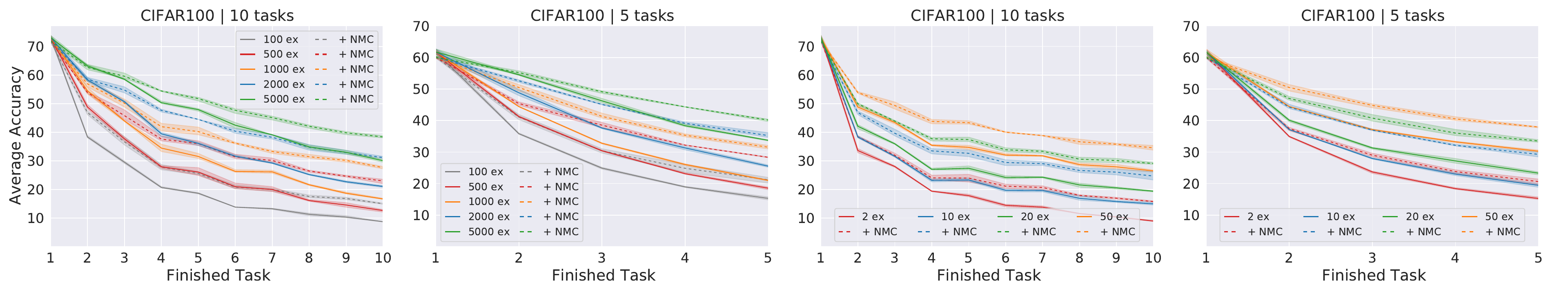}
   \caption{Average accuracy for FT and NMC over the course of continual learning with different memory budgets for CIFAR100. We present the results with fixed memory size on the left and with the growing memory on the right. NMC outperforms finetuning regardless of the memory selection algorithm.}
   \label{fig:number-of-exemplars}
\end{figure*}
\subsection{Impact of memory buffer size}
\label{sec:nmc-ablation}

NMC performs the classification using the class prototypes, and its performance will naturally be affected by the size of the exemplar set. Therefore, in this Section, we compare NMC and finetuning with varying numbers of exemplars and different types of memory. Specifically, we evaluate the networks with fixed-size memory where new data replaces some of the other exemplars throughout the training, and with growing memory where we keep a fixed number of exemplars for each class.

\begin{figure}[!t]
    \centering
    \includegraphics[width=0.49\linewidth]{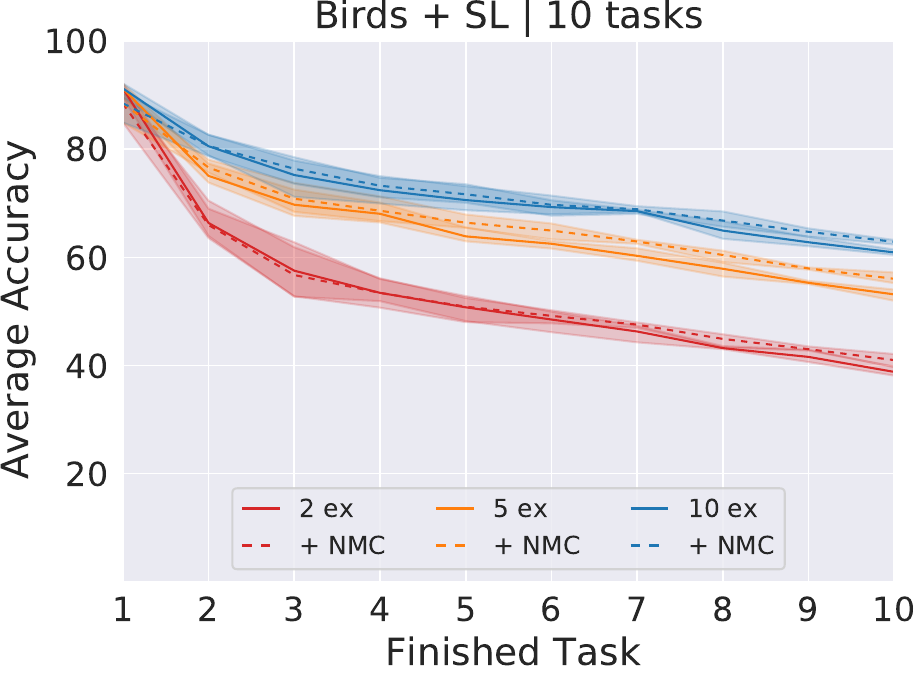}
    \hfill
    \includegraphics[width=0.49\linewidth]{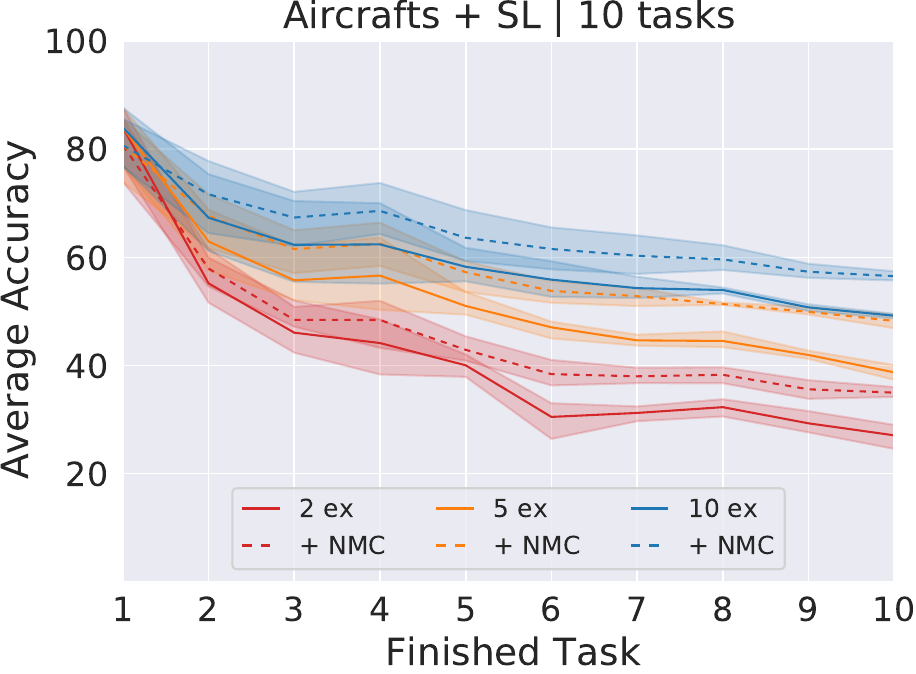}
   \caption{Average accuracy for FT and NMC on fine-grained datasets. NMC maintains its advantage over standard finetuning also when starting from the pretrained model.}
   \label{fig:number-of-exemplars-finegrained}
\end{figure}

We present the results for CIFAR100 in \cref{fig:number-of-exemplars}. Our experiments demonstrate that NMC robustly outperforms vanilla finetuning in final performance regardless of the size of the exemplar set. In \cref{fig:number-of-exemplars-finegrained} we present that the observations also hold for the setting of fine-grained classification when starting from pretrained model.

\subsection{NMC in joint incremental learning}
\label{sec:joint-nmc}

\begin{figure}[!t]
  \centering
  \includegraphics[width=0.49\linewidth]{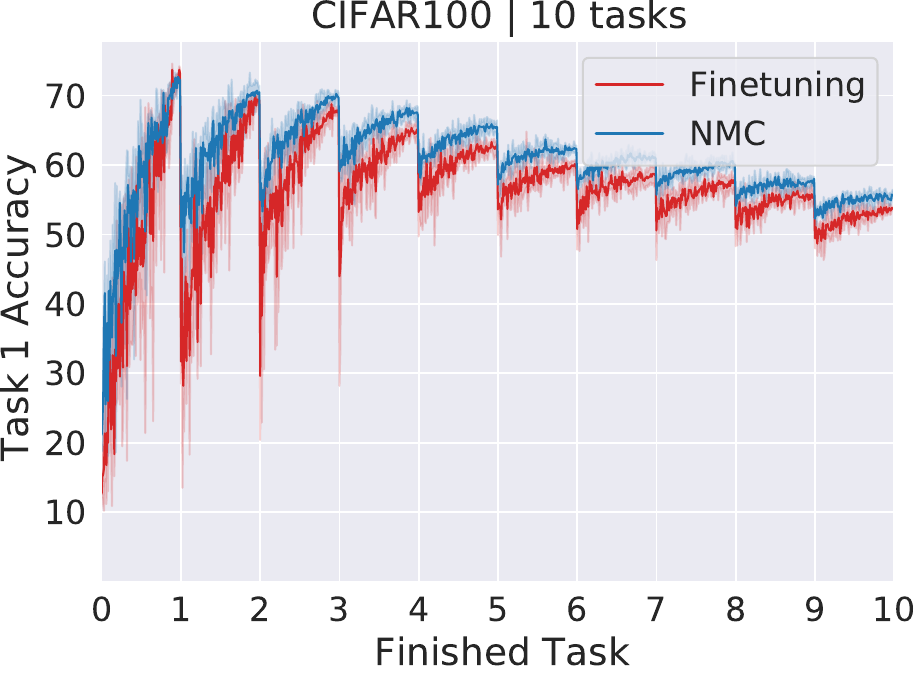}
  \hfill
  \includegraphics[width=0.49\linewidth]{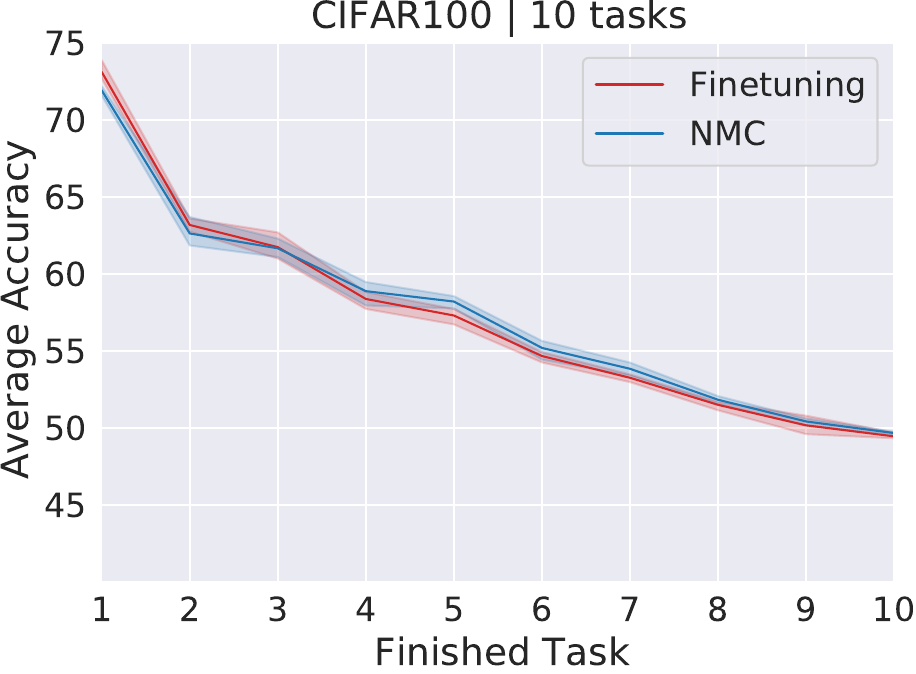}
  \caption{Joint incremental learning scenario. NMC maintains slight advantage over linear head when it comes to stability (left). However, the average accuracy over all the tasks is similar for both approaches (right).}
  \label{fig:joint-incr}
\end{figure}

Recent work~\cite{kamath2024expanding} showed that stability gap also happens in joint incremental learning. Therefore, it is valid for us to ask whether our previous findings on NMC's superior stability over the linear head also hold in this setting. To this end, we evaluate joint incremental learning models with standard head and NMC at \cref{fig:joint-incr} and examine their stability at the data from task 1, as well as average accuracy across all the tasks. Surprisingly, even in this scenario, NMC performs better than linear heads. This proves that non-parametric NMC improves learning stability and slightly helps with the stability gap even in scenarios with minimal changes in data distribution between the tasks.

\subsection{Task Recency Bias}
\label{sec:task-recency-bias}

In this Section, we explore task recency bias of finetuning and NMC.
To analyze this, we use the task confusion matrix, a variation of the classic confusion matrix, where the focus is on tasks rather than individual classes. In scenarios where each task contains multiple classes (e.g., 10 classes per task), the task-confusion matrix aggregates these predictions, providing insights into how well the model distinguishes between tasks rather than just individual classes. We denote the task confusion matrix as $C$ where $C_{ij}$ represents the number of instances from task $i$ predicted as task~$j$.

\begin{figure}
  \centering
  \begin{subfigure}{0.48\linewidth}
    \includegraphics[width=1.0\linewidth]{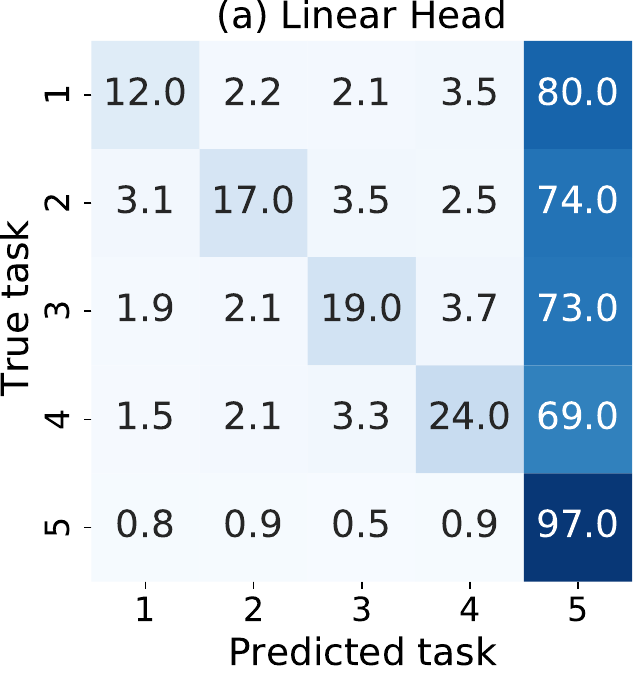}
    \label{fig:trb-a}
  \end{subfigure}
  \hfill
  \begin{subfigure}{0.48\linewidth}
    \includegraphics[width=1.0\linewidth]{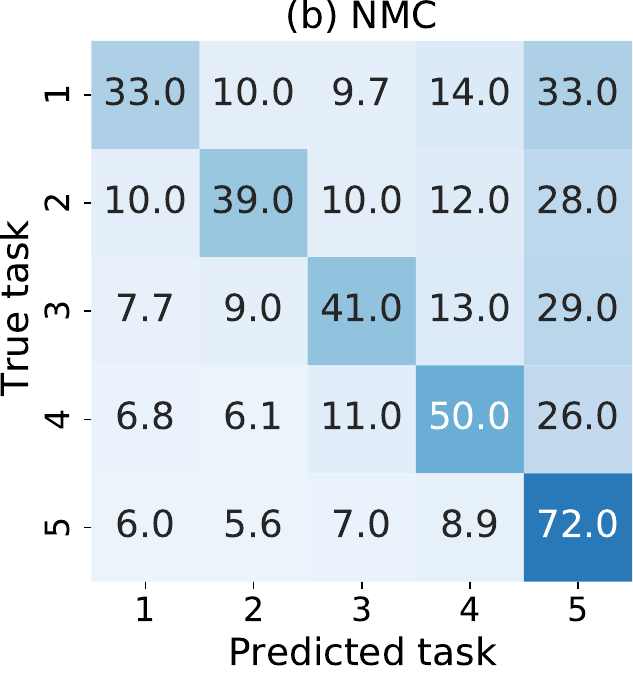}
    \label{fig:trb-b}
  \end{subfigure}
  \vspace{0.3cm}  
  \begin{subfigure}{0.48\linewidth}
    \includegraphics[width=1.0\linewidth]{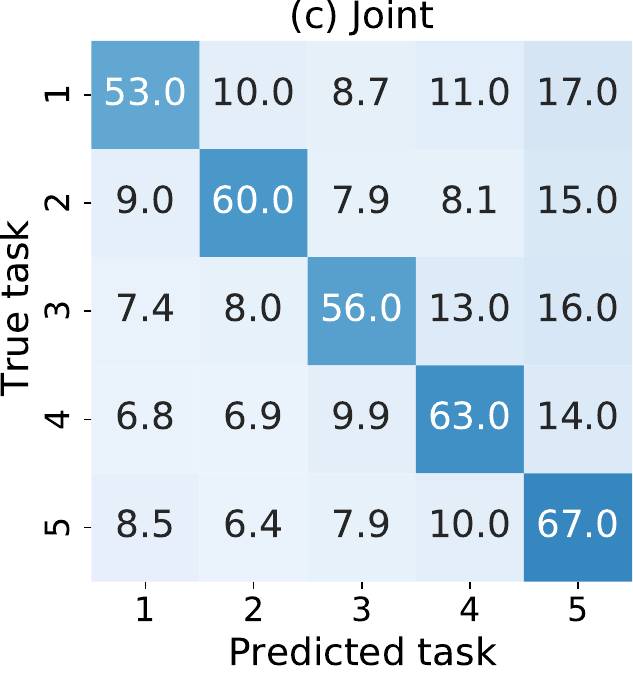}
    \label{fig:trb-c}
  \end{subfigure}
  \hfill
  \begin{subfigure}{0.48\linewidth}
    \includegraphics[width=1.0\linewidth]{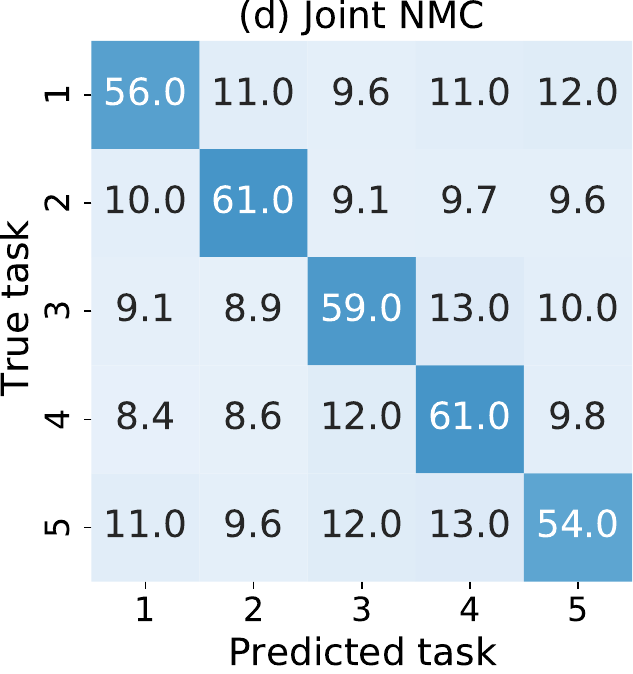}
    \label{fig:trb-d}
  \end{subfigure}
  \vspace{-0.3in}
  \caption{
  \textbf{Finetuning (a) is much more prone to task-recency bias than NMC (b). } Interestingly, we observe that in a joint incremental learning scenario, finetuning (c) still exhibits some bias towards the final task contrary to NMC (d). We present the task confusion matrix in \% (row-wise normalized). 
  }
  \label{fig:trb}
\end{figure}

In \cref{fig:trb} we present task confusion matrices after training on the final task. We observe severe task recency bias for finetuning with a linear head -- predictions of the classes from previous tasks are much less frequent than predictions from the last task. For NMC, however, the bias is much less evident and the task confusion matrix is closer to the ideal case of a diagonal matrix.

To continually evaluate the network's bias towards predicting the most recent task, we propose the \textbf{Latest Task Prediction Bias (LTB)} metric and define it as the mean of the values from matrix $C$ for tasks $1 \leq i \leq t-1$ mispredicted as $t$:
\begin{equation}
    \textbf{LTB}_{t,k} = \frac{1}{t-1} \sum_{i=1}^{t-1} C_{it}
    \label{eq:ltb}
\end{equation}
\noindent
where $k$ is the iteration index of task $t$. LTB measures how often the current task classes are mispredicted.

\begin{figure}[!t]
  \centering
  \begin{subfigure}{0.49\linewidth}
  \includegraphics[width=1.0\linewidth]{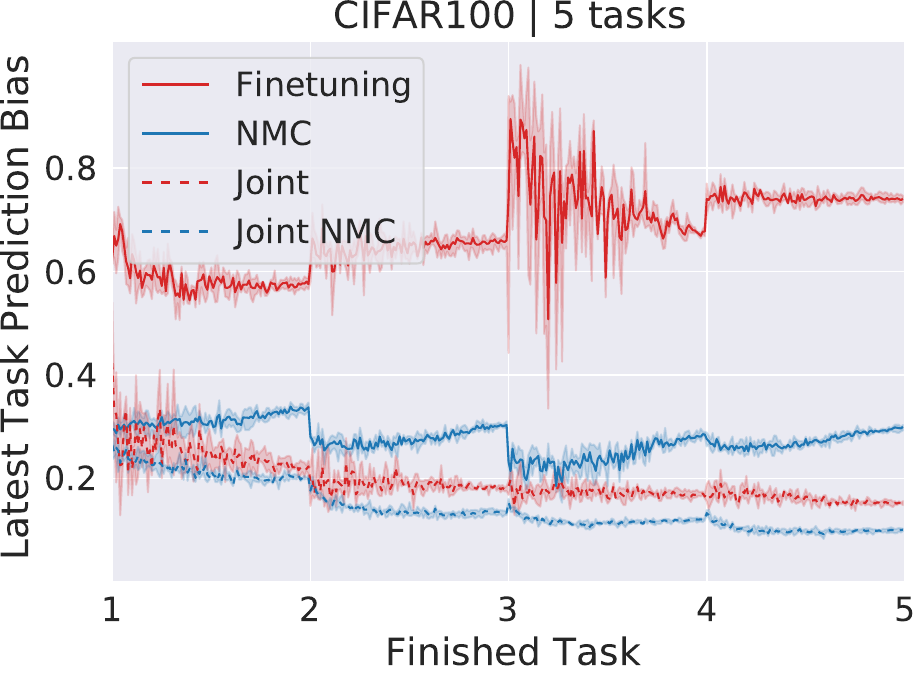}
  \end{subfigure}
  \centering
  \begin{subfigure}{0.49\linewidth}
  \includegraphics[width=1.0\linewidth]{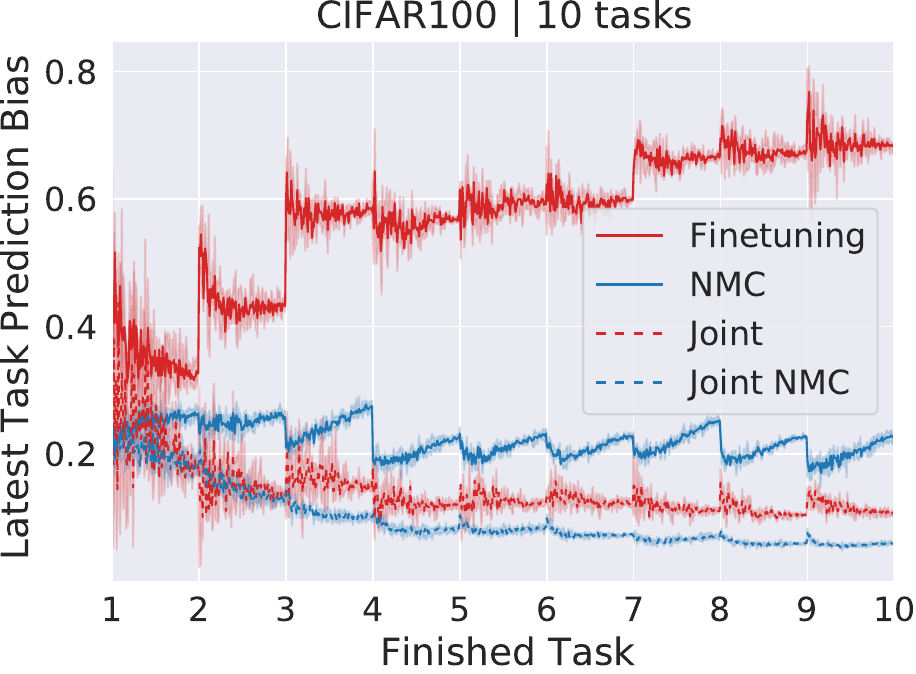}
  \end{subfigure}
   \caption{\textbf{NMC has significantly lower and more stable latest task prediction bias than linear heads.} LTB of NMC is relatively close to the results from joint incremental learning (where the learner has access to all the data from the previous task) which practically can be considered as a lower bound.}
   \label{fig:trb-metric}
\end{figure}

\cref{fig:trb-metric} shows that linear head's average \textbf{LTB} increases during training subsequent tasks, while NMC keeps it low and stable throughout the training.
This suggests that because the training set for the new task is significantly larger than the set of exemplars from previous tasks, linear heads tend to prioritize recent tasks over older ones (note that in the case of joint incremental training, LTB is low and stable - the bottom row of \cref{fig:trb-metric}). In contrast, NMC mitigates this issue by relying solely on embeddings from the backbone network, which allows NMC to maintain more balanced predictions across tasks, avoiding the recency bias seen in linear heads.

\subsection{Stability with classification head warm-up}

\begin{figure}[!t]
  \centering
  \begin{subfigure}{0.49\linewidth}
  \includegraphics[width=1.0\linewidth]{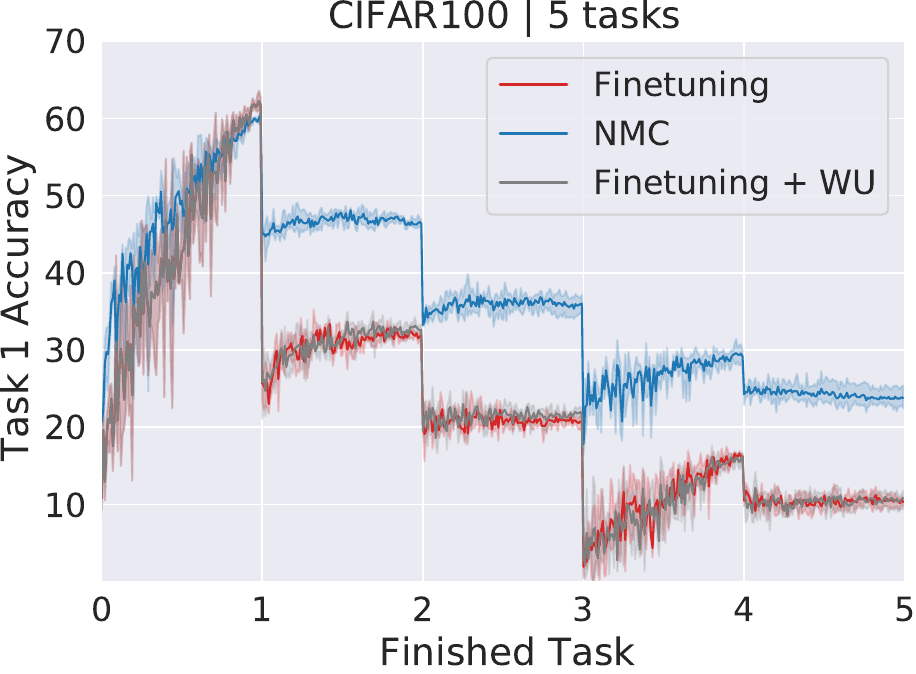}
  \end{subfigure}
  \centering
  \begin{subfigure}{0.49\linewidth}
  \includegraphics[width=1.0\linewidth]{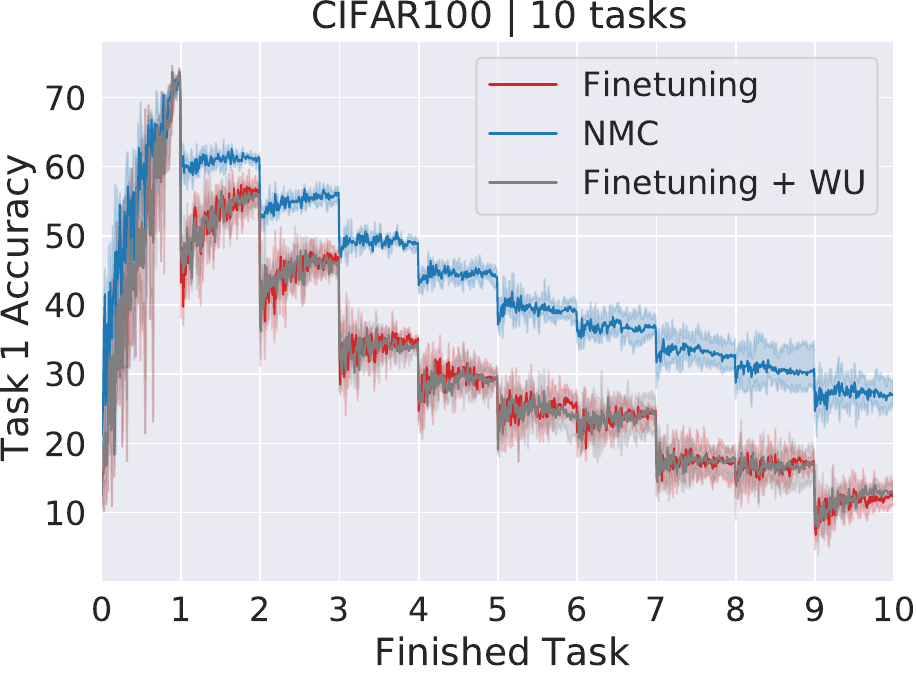}
  \end{subfigure}
   \caption{Warming up a new head before full finetuning does not improve the stability of finetuning. We present the accuracy on the first task. Warm-up training steps are not shown in the plot.}
   \label{fig:warmup}
   \vspace{-0.3cm}
\end{figure}

Work from Kumar et al.~\cite{kumar2022finetuning} suggests that a warm-up protocol for linear heads improves the downstream performance in transfer learning by first training new linear heads in isolation before finetuning the full network. Such head initialization reduces the initial error when learning a new task and therefore reduces the magnitude of the subsequent gradient updates, overall leading to fewer changes in the backbone network. The proposed warm-up scheme promises to improve learning stability, so motivated by the weak results of standard heads in our previous experiments we extend this protocol to CL to see if it can partially mitigate the performance difference between standard head and NMC. At the beginning of the new task, we first freeze the backbone and train only the linear head for a few epochs; then, we finetune both the backbone and head simultaneously. We present the results in \cref{fig:warmup}. We do not observe any benefits from employing the warm-up protocol in our CL scenario. However, we note that in the transfer learning setup from the original paper, the authors do not evaluate the performance on the source dataset. In our case, all previous tasks compose our source dataset and they are objects of consideration to evaluate the stability gap.


\section{Conclusions}

In this paper, we explore the stability gap phenomenon in continual learning, focusing on its origins within different components of neural networks: feature extractor and classification head. To disentangle the impact of each part of the model on the stability gap, we employ a non-parametric nearest-mean classifier. 
We compare the stability and performance of NMC and linear classification head on a comprehensive set of established continual learning benchmarks for general purpose and fine-grained image classification. To validate the robustness of our findings, we also vary the size of the memory buffer for the experiments and investigate joint training scenarios. We demonstrate that NMC not only improves the final performance but also significantly enhances stability throughout the training process both for randomly initialized and pre-trained networks. Finally, we show that NMC alleviates task-recency bias.

Our investigation suggests that the stability gap is predominantly caused by the linear classification head, rather than insufficient representations in the network backbone. We shed light on the underexplored topic of stability during continual learning and give a simple yet effective solution to mitigate the stability issues. We hope our study will encourage further exploration into the mechanisms underlying stability in continual learning and lead to the development of more robust and stable learning systems capable of adapting to dynamic data distributions.


\noindent\textbf{Limitations and future work}
Our work considers a simplified setting with exemplars for the sake of simplicity of the analysis. Extending the analysis to regularization based and exemplar-free methods remains an interesting direction for future work.

\noindent\textbf{Social Impact} 
Our study focuses on fundamental machine learning research and as such does not raise any specific ethical concerns. Nonetheless, we underscore the importance of the responsible application and deployment of machine learning algorithms in practice.

\noindent\textbf{Reproducibility}
All experiments conducted in this study leverage publicly available datasets and model checkpoints. To ensure the reproducibility of our results, we have provided the code for our work at \url{https://github.com/WojciechL02/stability-gap-nmc}.

\section*{Acknowledgments}
We gratefully acknowledge Poland's high-performance Infrastructure PL-Grid for providing computer facilities and support within computational grant no. PLG/2024/017385. This research was partially funded by National Science Centre, Poland, grant no: 2022/45/B/ST6/02817.

\clearpage

{\small
\bibliographystyle{ieee_fullname}
\bibliography{egbib}
}

\clearpage

\begin{appendix}

\twocolumn[\appendixhead]


\section{Task-Aware Accuracy}
\begin{figure*}[ht]
  \centering
    \includegraphics[width=1.0\linewidth]{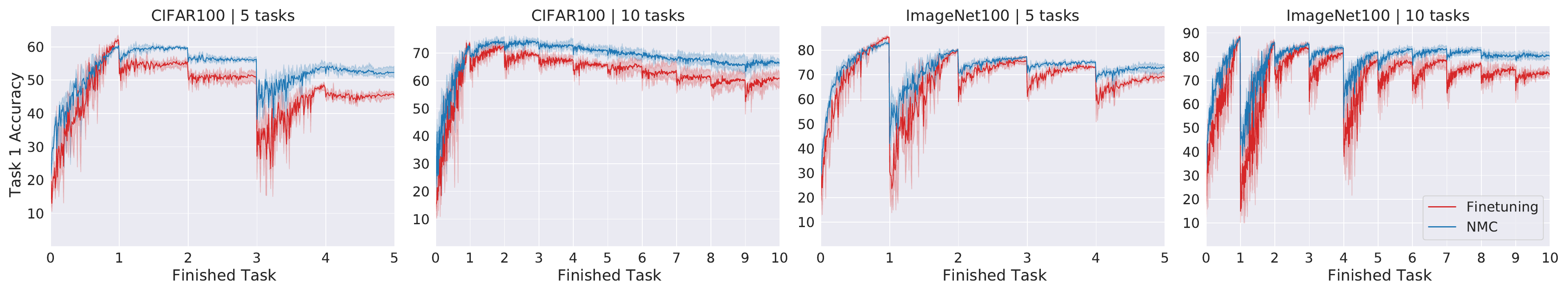}
  \caption{Task-aware results demonstrate that even when using a task-specific linear head, performance is lower than with NMC, reinforcing our previous claims.}
  \label{fig:taw-main}
\end{figure*}

\begin{figure*}[ht]
  \centering
    \includegraphics[width=1.0\linewidth]{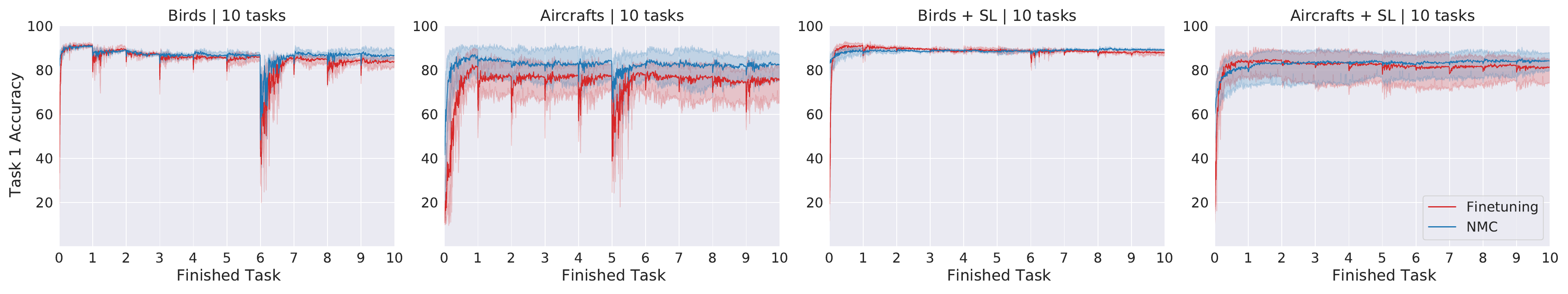}
  \caption{Task 1 task-aware accuracy during training on finegrained datasets.}
  \label{fig:taw-fin}
\end{figure*}

\begin{figure*}[ht]
  \centering
    \includegraphics[width=1.0\linewidth]{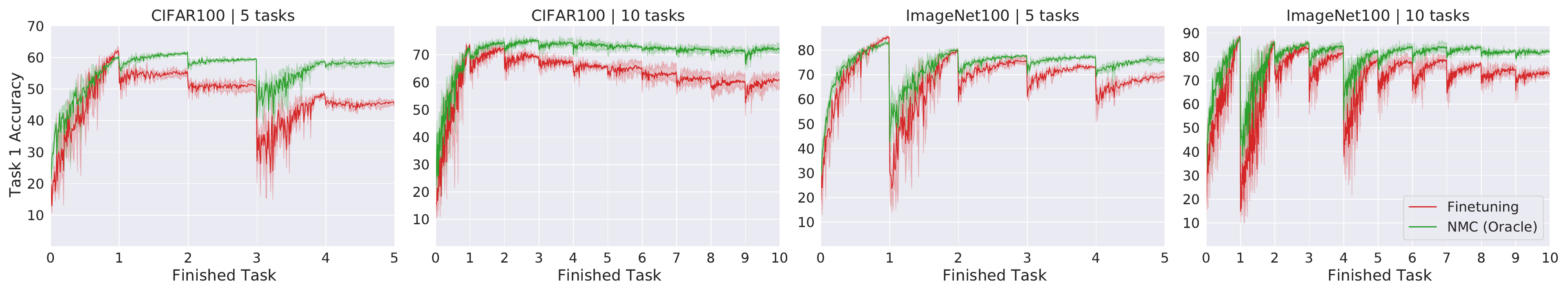}
  \caption{Oracle NMC has the best prototype estimates so it further improves TAw accuracy.}
  \label{fig:taw-oracle}
\end{figure*}

Task-aware accuracy is a metric used in continual learning to evaluate model performance when task identity is known during inference. Task-aware accuracy requires the model only to distinguish classes within-task, as opposed to task-agnostic accuracy, which requires both within-task class separation and correct task classification. Therefore, task-aware accuracy is considered to be an easier setting. In this section, we show task-aware results corresponding to experiments from~\cref{sec:disentangle} and~\cref{sec:realistic-scenarios}, and show that even when we evaluate task-specific linear head it performs worse than NMC, empowering our previous claims. The results can be seen in~\cref{fig:taw-main},~\cref{fig:taw-fin}, and~\cref{fig:taw-oracle}.

\section{Continual evaluation metrics}
\begin{figure*}[ht]
  \centering
    \includegraphics[width=1.0\linewidth]{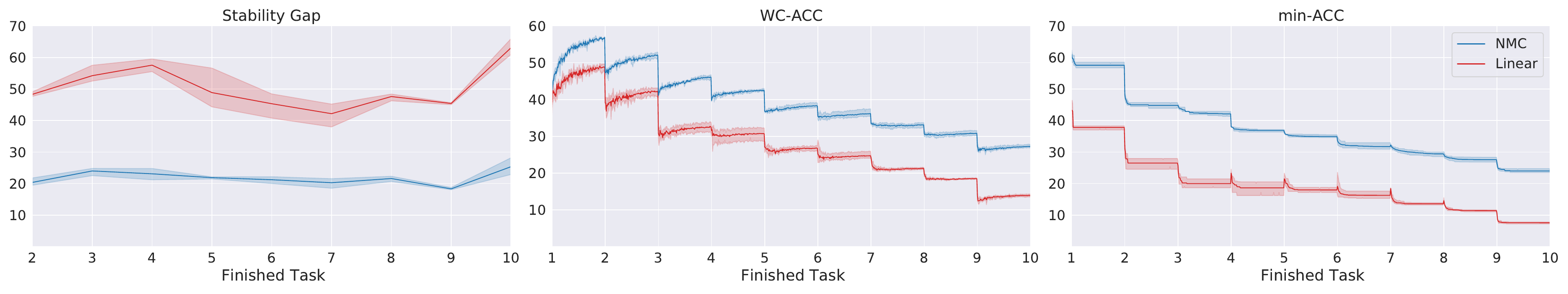}
  \caption{Fine-tuning on CIFAR100 (10 tasks). Continual evaluation metrics through full training of 10 tasks.}
  \label{fig:ceval-metrics}
\end{figure*}

The stability gap (SG) is a per-task metric, making it particularly useful for analyzing the dynamics of a model's behavior across individual tasks during sequential learning. Similarly, worst-case accuracy (WC-ACC) and average minimum accuracy (min-ACC) are metrics that can be evaluated on a per-iteration basis, providing further insights into the model's performance during training. Visualizing these metrics for all tasks on the plots is more informative than considering only the final scores, as it highlights how the model evolves and adapts over time. We present the plots of those metrics in~\cref{fig:ceval-metrics}.

\section{Other approaches}
\begin{figure*}[ht]
  \centering
    \includegraphics[width=1.0\linewidth]{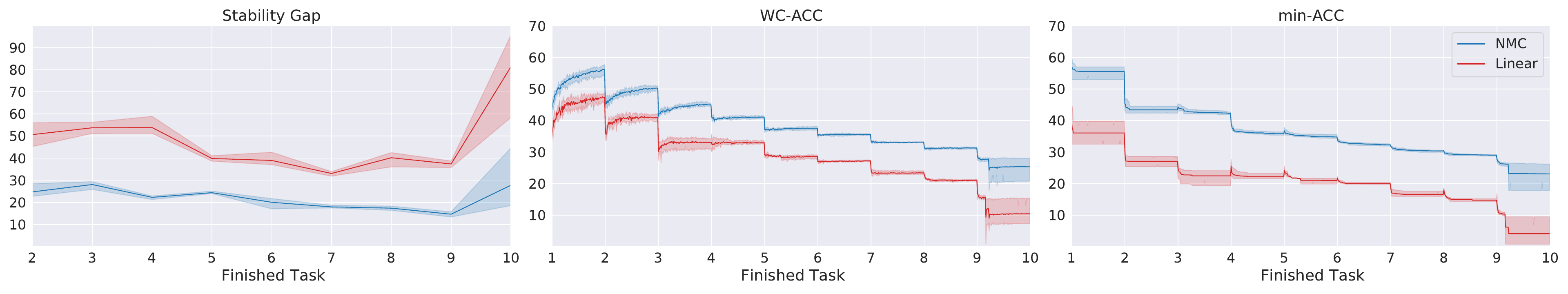}
  \caption{LwF on CIFAR100 (10 tasks). Continual evaluation metrics.}
  \label{fig:lwf-cont-ev}
\end{figure*}

\begin{figure*}[ht]
  \centering
    \includegraphics[width=1.0\linewidth]{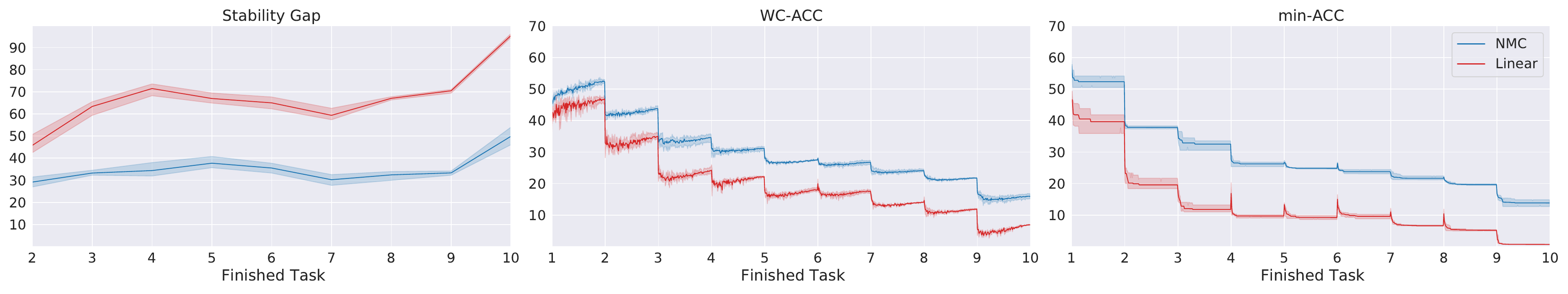}
  \caption{EWC on CIFAR100 (10 tasks). Continual evaluation metrics.}
  \label{fig:ewc-cont-ev}
\end{figure*}

\begin{figure*}[ht]
  \centering
    \includegraphics[width=1.0\linewidth]{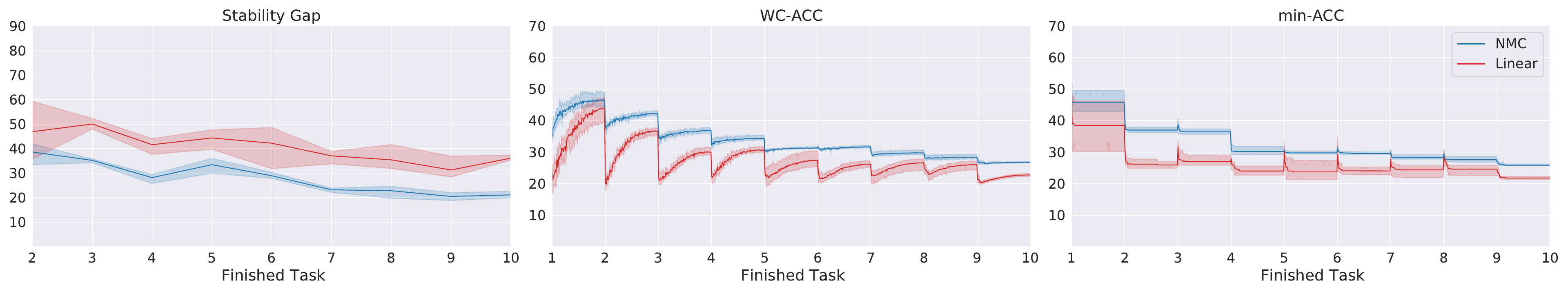}
  \caption{SS-IL on CIFAR100 (10 tasks). Continual evaluation metrics.}
  \label{fig:ssil-cont-ev}
\end{figure*}

We expand our investigation to regularization-based (LwF~\cite{li2017learning}, SS-IL~\cite{ahn2021ss}), and parameter-isolation (EWC~\cite{kirkpatrick2017overcoming}) methods. We use a constant memory buffer for all the methods (2000 exemplars). We also present continual evaluation metrics on the whole training on CIFAR100 (see~\cref{fig:lwf-cont-ev},~\cref{fig:ewc-cont-ev}, and~\cref{fig:ssil-cont-ev}).

\subsection{Task-Agnostic results}
\begin{figure*}[ht]
  \centering
    \includegraphics[width=1.0\linewidth]{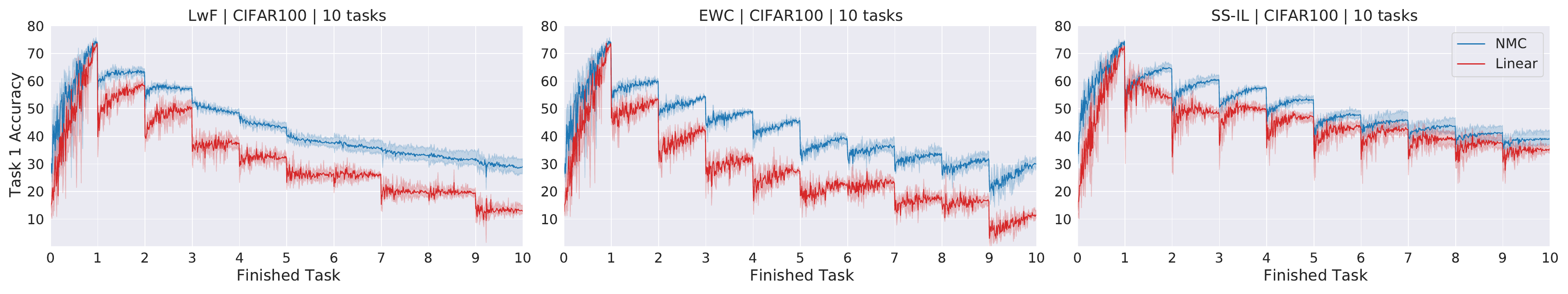}
  \caption{TAg accuracy. More advanced methods yield improved latent representations, while NMC further alleviates issues related to the classifier head. Observations from main experiments scale to other approaches.}
  \label{fig:other-appr}
\end{figure*}

More advanced methods, which usually perform better on CIL setups, provide better knowledge transfer while reducing forgetting and ultimately better representations, but still suffer from the linear multi-head classifier. Methods based on modified loss functions provide better latent representations, and NMC additionally mitigates the problems associated with the classifier head (see~\cref{tab:other-appr},~\cref{fig:other-appr}, and~\cref{fig:other-appr-ltb}).

\begin{table}[ht]
    \centering
    \caption{
    NMC improves the stability metrics and final accuracy of different CIL methods.
    }
    \scalebox{0.8}{
    \begin{tabular}{lcccc}
    \toprule
    & WC-ACC ($\uparrow$) & min-ACC ($\uparrow$) & SG($\downarrow$) & ACC ($\uparrow$) \\ \midrule
    & \multicolumn{4}{c}{CIFAR100/10} \\
    \cmidrule(lr){2-5}
    LwF                                          & 10.44\scriptsize{$\pm$4.32} & 4.12\scriptsize{$\pm$4.67} &  81.36\scriptsize{$\pm$15.08} & 21.55\scriptsize{$\pm$0.55}    \\
    +NMC                                         & \textbf{25.33}\scriptsize{$\pm$\textbf{4.00}} & \textbf{23.08}\scriptsize{$\pm$\textbf{4.60}} & \textbf{27.79}\scriptsize{$\pm$\textbf{8.83}} & \textbf{30.97}\scriptsize{$\pm$\textbf{0.36}}   \\
    \midrule
    EWC                                          & 6.92\scriptsize{$\pm$0.10} & 0.71\scriptsize{$\pm$0.04} & 95.38\scriptsize{$\pm$0.89} & 17.87\scriptsize{$\pm$0.27}   \\
    +NMC                                         & \textbf{15.95}\scriptsize{$\pm$\textbf{0.83}} & \textbf{13.87}\scriptsize{$\pm$\textbf{1.07}} & \textbf{49.88}\scriptsize{$\pm$\textbf{3.97}} & \textbf{25.84}\scriptsize{$\pm$\textbf{0.26}}  \\ \midrule
    SS-IL                                          &  22.72\scriptsize{$\pm$0.45} & 21.75\scriptsize{$\pm$0.45} & 36.09\scriptsize{$\pm$1.10} & \textbf{31.81}\scriptsize{$\pm$\textbf{0.39}}  \\
    +NMC                                         &  \textbf{26.74}\scriptsize{$\pm$\textbf{0.23}} & \textbf{25.83}\scriptsize{$\pm$\textbf{0.31}} & \textbf{21.11}\scriptsize{$\pm$\textbf{1.34}} & 31.05\scriptsize{$\pm$0.53}  \\
    \midrule
    & \multicolumn{4}{c}{ImageNet100/10} \\
    \cmidrule(lr){2-5}
    LwF                                          & 25.78\scriptsize{$\pm$0.36} & 19.66\scriptsize{$\pm$0.36} &  41.24\scriptsize{$\pm$1.06} & 31.36\scriptsize{$\pm$0.30}    \\
    +NMC                                         & \textbf{43.09}\scriptsize{$\pm$\textbf{0.13}} & \textbf{42.31}\scriptsize{$\pm$\textbf{0.08}} & \textbf{10.67}\scriptsize{$\pm$\textbf{0.42}} & \textbf{45.66}\scriptsize{$\pm$\textbf{0.46}}   \\
    \midrule
    EWC                                          & 17.9\scriptsize{$\pm$0.88} & 10.12\scriptsize{$\pm$0.89} & 68.87\scriptsize{$\pm$2.64} & 31.46\scriptsize{$\pm$0.39}   \\
    +NMC                                         & \textbf{36.01}\scriptsize{$\pm$\textbf{0.99}} & \textbf{30.17}\scriptsize{$\pm$\textbf{1.04}} & \textbf{34.46}\scriptsize{$\pm$\textbf{1.97}} & \textbf{44.58}\scriptsize{$\pm$\textbf{0.30}}  \\ \midrule
    SS-IL                                          &  37.42\scriptsize{$\pm$0.45} & 35.43\scriptsize{$\pm$0.81} & 21.53\scriptsize{$\pm$0.74} & 46.05\scriptsize{$\pm$0.67}  \\
    +NMC                                         &  \textbf{38.75}\scriptsize{$\pm$\textbf{0.53}} & \textbf{36.89}\scriptsize{$\pm$\textbf{0.61}} & \textbf{9.15}\scriptsize{$\pm$\textbf{0.94}} & \textbf{46.52}\scriptsize{$\pm$\textbf{0.42}}  \\
    \bottomrule
    \end{tabular}}
    \label{tab:other-appr}
\end{table}

\subsection{Latest-Task Prediction Bias}
\begin{figure*}[ht]
  \centering
    \includegraphics[width=1.0\linewidth]{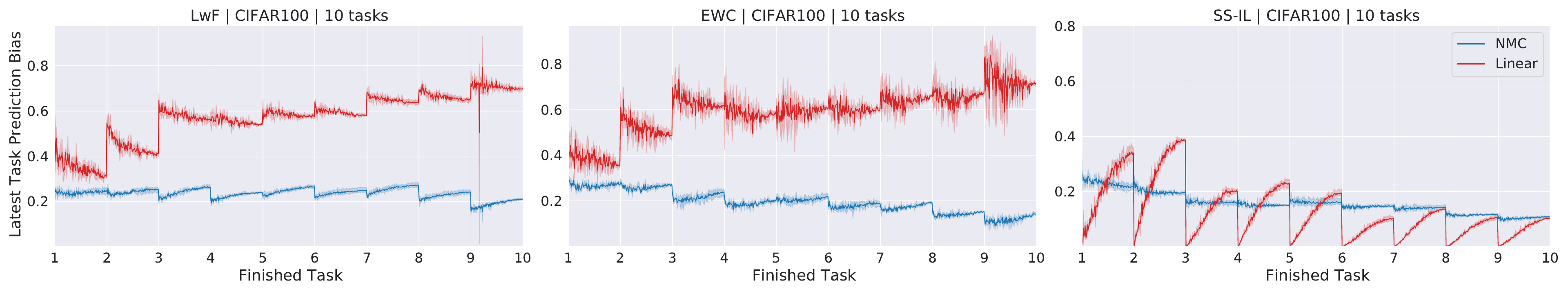}
  \caption{LwF and EWC suffer from latest task prediciton bias, but it can be reduced with NMC. SS-IL has internal mechanism to mitigate task-recency bias, but still its LTB is comparable to NMC's.}
  \label{fig:other-appr-ltb}
\end{figure*}

Experiment analogous to the one in~\cref{fig:trb-metric}. We observe that NMC reduces the LTB even in approaches that try to improve the linear heads of previous tasks, e.g. by knowledge distillation or other methods, more than simple finetuning with exemplars (see results in~\cref{fig:other-appr-ltb}).

\begin{figure*}[ht]
  \centering
    \includegraphics[width=1.0\linewidth]{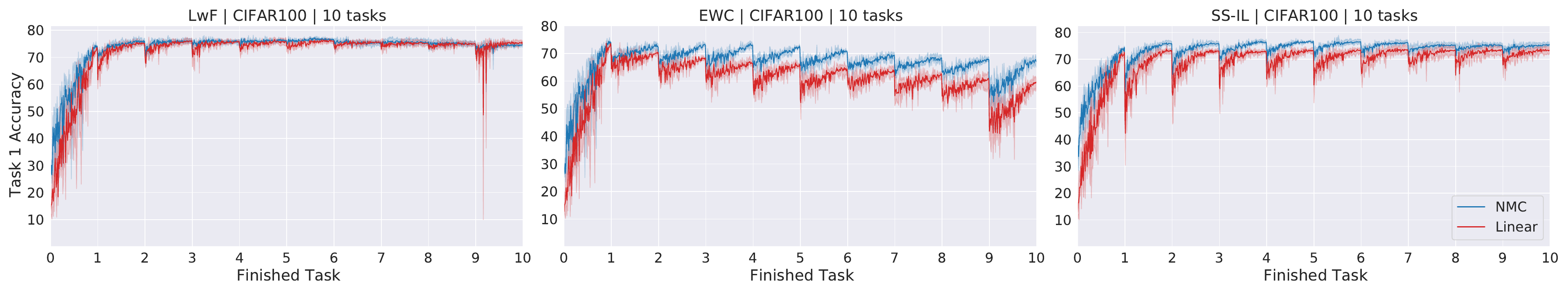}
  \caption{For completeness we also present TAw accuracy scores.}
  \label{fig:other-appr-taw}
\end{figure*}

\section{Different CNN architectures}
To further investigate the influence of the non-parametric NMC classifier, in addition to experiments with ResNet18, we test it with other standard convolutional neural networks. We evaluate finetuning with constant memory (2000 exemplars) on linear multi-head and NMC with MobileNetV2~\cite{mobilenetv2}, ResNet50~\cite{he2016resnet}, EfficientNet-B4~\cite{tan2020efficientnetrethinkingmodelscaling}, and VGG11~\cite{simonyan2015deepconvolutionalnetworkslargescale} as backbones. We use CIFAR100 split into 10 equally sized tasks and train the networks as described in~\cref{sec:incr-learn}. We still notice that using NMC constantly improves the results, regardless of the architecture we use.

\begin{table}[ht]
    \centering
    \caption{
    CIFAR100 split into 10 disjoint tasks.
    }
    \scalebox{0.75}{
    \begin{tabular}{lcccc}
    \toprule
    Network & WC-ACC ($\uparrow$) & min-ACC ($\uparrow$) & SG($\downarrow$) & ACC ($\uparrow$) \\ \midrule
    ResNet18                                       & 13.94\scriptsize{$\pm$0.39} & 7.58\scriptsize{$\pm$0.32} & 63.04\scriptsize{$\pm$2.57} & 21.04\scriptsize{$\pm$0.31}   \\
    +NMC                                        & \textbf{27.27}\scriptsize{$\pm$\textbf{0.50}} & \textbf{24.01}\scriptsize{$\pm$\textbf{0.65}} & \textbf{25.35}\scriptsize{$\pm$\textbf{2.66}} & \textbf{31.14}\scriptsize{$\pm$\textbf{0.38}}  \\ \midrule
    ResNet50                                        & 7.43\scriptsize{$\pm$1.79} & 1.51\scriptsize{$\pm$2.12} & 89.23\scriptsize{$\pm$15.16} & 14.92\scriptsize{$\pm$0.62}   \\
    +NMC                                        & \textbf{11.49}\scriptsize{$\pm$\textbf{8.43}} & \textbf{7.26}\scriptsize{$\pm$\textbf{9.08}} & \textbf{69.76}\scriptsize{$\pm$\textbf{29.29}} & \textbf{23.36}\scriptsize{$\pm$\textbf{1.31}}  \\ \midrule
    VGG11                                        &  16.58\scriptsize{$\pm$0.40} & 10.61\scriptsize{$\pm$0.45} & 57.44\scriptsize{$\pm$0.71} & 25.14\scriptsize{$\pm$0.23}  \\
    +NMC                                        &  \textbf{24.75}\scriptsize{$\pm$\textbf{0.61}} & \textbf{20.15}\scriptsize{$\pm$\textbf{0.75}} & \textbf{31.31}\scriptsize{$\pm$\textbf{2.13}} & \textbf{28.81}\scriptsize{$\pm$\textbf{0.39}}  \\ \midrule
    MobileNetV2                                        &  11.49\scriptsize{$\pm$0.76} & 5.56\scriptsize{$\pm$0.94} & 65.56\scriptsize{$\pm$7.53} & 17.56\scriptsize{$\pm$0.91}  \\
    +NMC                                        &  \textbf{22.72}\scriptsize{$\pm$\textbf{0.53}} & \textbf{19.43}\scriptsize{$\pm$\textbf{0.59}} & \textbf{24.93}\scriptsize{$\pm$\textbf{0.96}} & \textbf{25.44}\scriptsize{$\pm$\textbf{0.55}}  \\ \midrule
    EfficientNet-B4                                       &  6.94\scriptsize{$\pm$0.70} & 0.43\scriptsize{$\pm$0.68} & 97.65\scriptsize{$\pm$3.50} & 17.79\scriptsize{$\pm$2.16}  \\
    +NMC                                       &  \textbf{14.02}\scriptsize{$\pm$\textbf{5.71}} & \textbf{9.6}\scriptsize{$\pm$\textbf{6.23}} & \textbf{63.77}\scriptsize{$\pm$\textbf{23.88}} & \textbf{27.10}\scriptsize{$\pm$\textbf{0.50}}  \\
    \bottomrule
    \end{tabular}}
    \label{tab:apx-main-results}
\end{table}

\end{appendix}

\end{document}